\documentclass{article}
\usepackage{microtype}
\usepackage{graphicx}
\usepackage{subfigure}
\usepackage{booktabs}

\usepackage{hyperref}

\usepackage[accepted]{icml2025}

\usepackage{amsmath}
\usepackage{amssymb}
\usepackage{mathtools}
\usepackage{amsthm}
\usepackage[algo2e]{algorithm2e} 
\SetKwInput{KwIn}{Input}
\SetKwInput{KwOut}{Output}

\usepackage[capitalize,noabbrev]{cleveref}

\theoremstyle{plain}

\theoremstyle{definition}

\theoremstyle{remark}

\DeclareMathOperator*{\argmax}{arg\,max}

\usepackage[textsize=tiny]{todonotes}

\begin{document}

\twocolumn[
\icmltitle{Visual Language Models as Zero-Shot Deepfake Detectors}

\begin{icmlauthorlist}
    \icmlauthor{Viacheslav Pirogov}{sumsub}
\end{icmlauthorlist}

\icmlaffiliation{sumsub}{Sumsub, Berlin, Germany}

\icmlcorrespondingauthor{Viacheslav Pirogov}{slava.pirogov@sumsub.com}

\icmlkeywords{Deepfake Detection, Visual Language Models, Zero-Shot Learning}

\vskip 0.3in
]
\printAffiliationsAndNotice{}

\begin{abstract}
    The contemporary phenomenon of deepfakes, utilizing GAN or diffusion models for face swapping, presents a substantial and evolving threat in digital media, identity verification, and a multitude of other systems.
    The majority of existing methods for detecting deepfakes rely on training specialized classifiers to distinguish between genuine and manipulated images, focusing only on the image domain without incorporating any auxiliary tasks that could enhance robustness. \\
    In this paper, inspired by the zero‑shot capabilities of Vision–Language Models, we propose a novel VLM‑based approach to image classification and then evaluate it for deepfake detection. Specifically, we utilize a new high‑quality deepfake dataset comprising 60,000 images, on which our zero‑shot models demonstrate superior performance to almost all existing methods. Subsequently, we compare the performance of the best-performing architecture, InstructBLIP, on the popular deepfake dataset DFDC-P against traditional methods in two scenarios: zero-shot and in-domain fine-tuning. Our results demonstrate the superiority of VLMs over traditional classifiers.
\end{abstract}

\section{Introduction}
\label{sec:intro}
Deepfakes have evolved from a technological curiosity to an everyday reality. Just a few years ago, producing a high-quality, AI-generated image required specialized knowledge, powerful hardware and strong technical skills, so only a handful of enthusiasts could do it. However, thanks to the growing popularity of AI and the democratization of computing resources such as Google Colab \cite{GoogleColab}, as well as dozens of ready-made tools \cite{ComfyUI, openai_chatgpt_2024, Veo3}, almost anyone with a laptop, or even a smartphone, can now create deepfakes for free or for the price of a small subscription. Most of this AI-generated content appears on social media platforms, where people use it purely for entertainment. However, deepfakes also pose serious dangers, particularly with regard to Know Your Customer (KYC) onboarding and liveness verification. According to the Sumsub report \cite{sumsub_report}, the number of detected deepfakes increased fourfold between 2023 and 2024, accounting for 7\% of all fraud attempts.

Many studies have emphasized that the problem of detecting deepfakes remains significant and unsolved \cite{le2024sok, liu2024evolving, heidari2024deepfake, le2023facial}. One primary challenge is the absence of a comprehensive dataset that covers all types of deepfake encountered in real life. In addition, existing detectors are fragile: even simple post‑processing such as noise \cite{haliassos2021lipsdontliegeneralisable, jiang2020deeperforensics10largescaledatasetrealworld} or compression \cite{le2023qualityagnosticdeepfakedetectionintramodel, le2021addfrequencyattentionmultiview} can break them. While these models achieve high accuracy on their evaluation datasets, they often fail to generalize to new types of deepfake, and not ready for real-world applications, as demonstrated in our \textit{parallel} work \cite{pirogov2025evaluatingdeepfakedetectorswild}.

A truly practical solution would offer a robust zero‑shot or few‑shot method that can effectively detect previously unseen deepfakes. Modern large language models (LLMs) \cite{fraser_chatgpt_2023, geminiteam2024geminifamilyhighlycapable, deepseekai2025deepseekr1incentivizingreasoningcapability} are popular precisely because they exhibit strong zero‑shot and generalization abilities. Visual Language Models (VLMs) with instruction tuning \cite{dai2023instructblipgeneralpurposevisionlanguagemodels, liu2023visualinstructiontuning, laurençon2024mattersbuildingvisionlanguagemodels} demonstrate comparable potential. In natural‑language processing, most downstream systems now build on huge pretrained backbones and adapt them with lightweight methods. Therefore, a similar trend could be expected in computer vision, with VLMs fitting best in the role of such a backbone.

Previous studies have demonstrated the potential of both open-source and closed-source Visual Language Models (VLMs) in the task of deepfake detection \cite{chang2023antifakeprompt, zhang2024commonsensereasoningdeep, li2024fakebenchuncoverachillesheels, shi2024shieldevaluationbenchmark, jia2024chatgptdetectdeepfakesstudy}. However, these studies have not fully explored the pure zero-shot capabilities of VLMs and have not focused on integrating such models into real-life systems, such as liveness checks and verifications.

In this paper, we introduce a VLM‑based image‑classification framework and evaluate it on deepfake detection. 
Using a new high‑quality deepfake dataset, we benchmark state‑of‑the‑art detectors together with open and closed source VLMs in zero‑shot and few‑shot settings, showing that VLMs achieve the best out‑of‑distribution performance.
Furthermore, we demonstrate that, when the data are in-distribution, simple language fine-tuning enables the leading VLM to achieve near-perfect performance; we illustrate this using the widely adopted deepfake dataset DFDC-P \cite{dolhansky2020deepfakedetectionchallengedfdc}.

\section{Related work}

This section provides an overview of the techniques used to generate deepfakes in the modern world, with a particular focus on face swapping between two individuals \ref{sec:deepfake_gen} and methods for detecting such deepfakes \ref{sec:deepfake_det}. Subsequently, we present the current state of Visual Language Models (VLMs), with a particular focus on the VLMs that we have utilized \ref{sec:VLMs}. In the last subsection \ref{sec:VLMs_detection}, we offer a comprehensive overview of existing research that employs VLMs as deepfake detectors.

\subsection{Deepfake Generation}
\label{sec:deepfake_gen}

Face swapping approaches can be categorized based on the number of images required for the source and target faces. These approaches range from using large datasets, commonly referred to as "facesets", to employing few-shot or one-shot methods. The most powerful method utilizing large datasets is DeepFaceLab \cite{perov2020deepfacelab}. In contrast, few-shot or one-shot methods \cite{10.1145/3394171.3413630, nirkin2019fsgan, nirkin2022fsganv2, li2019faceshifter, InsightFace} are the most popular today within the community \cite{roop, roop_SD, roop_unleashed, Fooocus_inswapper} and among fraudsters due to their ease of use and low resource requirements.

\textbf{SimSwap}: An Efficient Framework For High Fidelity Face Swapping is a state-of-the-art (SOTA) open-source model for high-fidelity one-shot face swapping, requiring only one image each from source and target. This model supports relatively high resolutions with two variants: one at 224x224 pixels, trained on the VGGFace2 \cite{DBLP:vggface} dataset, and other at 512x512, trained on enhanced VGGFace2-HQ dataset. SimSwap utilizes an adversarially trained encoder-decoder architecture, augmented by an Identity Injection Module, which separates image attributes and identity, thereby enabling the transfer of only the attributes. To this end, a face recognition network \cite{DBLP:ArcFace} is employed to extract an identity embedding, which is then integrated via Adaptive Instance Normalization (AdaIN) \cite{DBLP:AdaIN}. The model is trained in a GAN style \cite{DBLP:GAN, DBLP:FUNIT, DBLP:BigGAN, DBLP:WGAN-GP, DBLP:styleGAN, DBLP:pix2pix} to ensure the realism of generated images, with the objective of achieving an indistinguishable result from that of the original.

\subsection{Deepfake Detection}
\label{sec:deepfake_det}

For the last five years, researchers have been actively working on deepfake detection methods. They usually propose new datasets that better represent real-life scenarios than previous ones \cite{rossler2019faceforensics++, dolhansky2020deepfakedetectionchallengedfdc, li2020celebdflargescalechallengingdataset, jiang2020deeperforensics10largescaledatasetrealworld, shiohara2022detecting} or introduce new analytical approaches such as novel architectures (\cite{zhao2021multi, wang2022m2tr, sun2021dualcontrastivelearninggeneral}), frequency-based methods \cite{le2021addfrequencyattentionmultiview, qian2020thinkingfrequencyfaceforgery, 10.1007/978-3-031-19830-4_27}, spatial techniques \cite{le2023qualityagnosticdeepfakedetectionintramodel, nguyen2018capsuleforensicsusingcapsulenetworks, Tariq_2021}, and many other approaches \cite{9878441, dong2023implicit, sun2024generalvisuallinguisticfaceforgery, chen2021localrelationlearningface, Sun_Liu_Ye_Gao_Liu_Shao_Ji_2021}.

The first significant work in deepfake area was \textbf{FaceForensics++} \cite{rossler2019faceforensics++}, which presented a dataset of over 1.8 million images from 1000 YouTube videos, accompanied by a simple detection model based on XceptionNet \cite{chollet2017xception}. Building on this, the authors of \textbf{MAT} \cite{zhao2021multi} proposed moving from a simple CNN model to a multi-attention network, inspired by the popularity of Visual Transformers \cite{dosovitskiy2020image, vaswani2017attention}. Similarly, the authors of \textbf{M2TR} \cite{wang2022m2tr} employed a frequency filter \cite{ricker2022towards} with a 2D Fast Fourier Transform to enhance detection. Another innovative approach is \textbf{RECCE} \cite{9878441}, which employs metric-learning loss and reconstruction learning \cite{wertheimer2021few} to improve upon previous methods. One of the most effective and robust methods is \textbf{SBI} \cite{shiohara2022detecting}, which presents a unique dataset generated by blending pseudo-source and target images derived from individual pristine images.

\subsection{Visual Language Models}
\label{sec:VLMs}

The first significant work in the modern state of Visual Language Models (VLMs) field was Flamingo, introduced in \cite{alayrac2022flamingovisuallanguagemodel}. The authors proposed utilizing pre-trained and frozen during fine-tuning Vision Encoder and Language Model (LM), connected by a Perceiver Resampler \cite{jaegle2021perceivergeneralperceptioniterative}, which processes varying-size large feature maps and outputs few visual tokens. These tokens are then fed through gated cross attention and trained with language modelling loss.
Another notable initial work was CoCa \cite{yu2022cocacontrastivecaptionersimagetext}, where the authors combined contrastive pre-training and language modelling into a single model. 
This was achieved by passing image and text pairs to the corresponding encoders, where a contrastive loss was calculated on the CLS tokens. Subsequently, a text decoder with cross-attention on image features was employed to calculate the language modelling loss.

One of the most significant open-source VLM families is the LAVIS family \cite{salesforce_lavis_2024}. The initial work "BLIP: Bootstrapping Language-Image Pre-training for Unified Vision-Language Understanding and Generation" was introduced in \cite{li2022blipbootstrappinglanguageimagepretraining}. The authors proposed an encoder-decoder architecture trained on three tasks: contrastive image-text pairing \cite{DBLP:journals/corr/abs-2103-00020}, image-text matching with cross-attention, and language modeling loss. A crucial part of this work was dataset bootstrapping with synthetic captions.
The continuation of this work is BLIP-2 \cite{li2023blip2bootstrappinglanguageimagepretraining}, which aimed to combine a pretrained and frozen image encoder and LLM with a lightweight module named Q-Former, containing a few self and cross-attention layers. Inspired by zero-shot capabilities of instruction tuning \cite{openai_chatgpt_2024}, the authors presented InstructBLIP \cite{dai2023instructblipgeneralpurposevisionlanguagemodels}, which involved the collection of a new instruction dataset and the fine-tuning of the Q-Former and frozen LLM, resulting in improved overall performance.

In 2023-2024, VLMs diverged into two main approaches: training large models on extensive datasets with different tasks \cite{alayrac2022flamingovisuallanguagemodel, li2022blipbootstrappinglanguageimagepretraining, yu2022cocacontrastivecaptionersimagetext}, and using a small connector between frozen Vision Encoders and LLMs \cite{li2023blip2bootstrappinglanguageimagepretraining, dai2023instructblipgeneralpurposevisionlanguagemodels, liu2023visualinstructiontuning, laurençon2024mattersbuildingvisionlanguagemodels}. In 2023, the authors of FROMAGe \cite{koh2023groundinglanguagemodelsimages} demonstrated that connecting a pretrained Visual Encoder and LLM could be achieved with just three linear layers as a projection and the addition of a special image token, with training for only one day on a single GPU. This straightforward and cost-effective approach outperformed many previous methods that had been trained on multiple GPUs over extended periods.

Inspired by FROMAGe \cite{koh2023groundinglanguagemodelsimages} and instruction tuning \cite{openai_chatgpt_2024}, LLaVA proposed in \cite{liu2023visualinstructiontuning}, with a creation of a new instruction image-language dataset. The authors used ChatGPT \cite{openai_chatgpt_2024} to generate captions and questions in a chatbot format without seeing the picture. Following 150 hours of training with FROMAGe-like architecture, the model became SOTA in many benchmarks, even surpassing closed GPT-4V \cite{openai2024gpt4technicalreport}. Subsequently, the authors released LLaVA 1.5 and 1.6 (LLaVA-NeXT) \cite{liu2024improvedbaselinesvisualinstruction}, with minor improvements in training data and trained models using other pre-trained LLMs. One of the most recent models developed by Hugging Face, called Idefics2 \cite{laurençon2024mattersbuildingvisionlanguagemodels}, is very similar to LLaVA, with Modality Projection and Pooling as connectors and its own instruction dataset.

\subsection{VLMs in deepfake detection}
\label{sec:VLMs_detection}
Researchers have already shown some potential of Visual Language Models (VLMs) in the deepfake detection task. Nevertheless, the full generalizability of these models in zero-shot or few-shot setups has not yet been demonstrated. In this subsection, we review existing methods, some of which focus on fine-tuning \cite{chang2023antifakeprompt}, while others reformulate the classification task into reasoning or Visual Question Answering (VQA) tasks \cite{zhang2024commonsensereasoningdeep, li2024fakebenchuncoverachillesheels}. A few studies also explore how to employ closed-source VLMs such as GPT-4V \cite{openai2024gpt4technicalreport} and Gemini \cite{geminiteam2024geminifamilyhighlycapable} \cite{shi2024shieldevaluationbenchmark, jia2024chatgptdetectdeepfakesstudy}.

The initial work, that utilized a Visual Language Model for deepfake detection is AntifakePrompt \cite{chang2023antifakeprompt}. The authors proposed to formulate deepfake detection as a Visual Question Answering (VQA) problem and tuning soft prompts for InstructBLIP \cite{dai2023instructblipgeneralpurposevisionlanguagemodels} to distinguish whether a query image is real or fake. They trained InstructBLIP on a real dataset sampled from MSCOCO \cite{lin2015microsoftcococommonobjects} and created their own fake dataset containing entirely or partly generated images, various types of adversarial attacks, and a small part of the Deeperforensics dataset \cite{jiang2020deeperforensics10largescaledatasetrealworld}. However, the authors did not focus on zero-shot capabilities and mainly addressed binary classification with a 0/1 prediction, which limits the ability to adjust the threshold, an important aspect in real-world applications.

The paper "Common Sense Reasoning for Deepfake Detection" \cite{zhang2024commonsensereasoningdeep} proposes the Deepfake Detection VQA (DD-VQA) task, which extends the domain of deepfake detection from conventional binary classification to a VQA task. Similarly, "FakeBench" \cite{li2024fakebenchuncoverachillesheels} presents a small image-level fake dataset for the evaluation of not only the classification accuracy of VLMs, but also their reasoning regarding the authenticity of images.

Two studies evaluated GPT-4V \cite{openai2024gpt4technicalreport} and Gemini \cite{geminiteam2024geminifamilyhighlycapable} for deepfake classification tasks. The first, "SHIELD" \cite{shi2024shieldevaluationbenchmark}, qualitatively evaluated various prompt techniques, ranging from simple questions like "Is it a deepfake?" to applying Multi-Attribute Chain of Thought (MA-COT) \cite{wei2023chainofthoughtpromptingelicitsreasoning, wu2023rolechainofthoughtcomplexvisionlanguage}. The second study, entitled "Can ChatGPT Detect DeepFakes?" \cite{jia2024chatgptdetectdeepfakesstudy}, quantitatively assessed GPT-4V and Gemini 1.0 in zero-shot setups with different prompts for deepfake detection. The prompts employed ranged from simple questions, such as "Tell me if this is an AI-generated image?" to more complex ones, including "Tell me the probability of this image being AI-generated." Tests conducted on a simple deepfake dataset with augmentations demonstrated the potential of VLMs to return a probability score, with high AUC scores achieved.

\section{Methodology}
In this section, we propose a new method for deepfake classification using VLMs \ref{sec:token_classification}, extend this method to multi-class tasks and multi-token answers \ref{sec:extenstion}, and discuss the crucial part of prompt engineering \ref{sec:prompt}, which is particularly important for closed VLMs.

\subsection{Classification}
\label{sec:token_classification}

\textbf{Roadmap.}  
\emph{We (i) recap the vanilla “arg-max” VQA baseline, (ii) expose its shortcomings, (iii) derive our probabilistic reformulation, (iv) give a worked numeric example, and (v) summarize how the new score feeds standard biometric metrics.}

A straightforward method for classifying an image with a VLM is to provide the model with an image and ask a question regarding the image's label, as has been employed in previous methods \cite{chang2023antifakeprompt, li2024fakebenchuncoverachillesheels, zhang2024commonsensereasoningdeep, shi2024shieldevaluationbenchmark, jia2024chatgptdetectdeepfakesstudy}. This might involve simple questions like "Tell me if this is an AI-generated image. Answer yes or no." or more complex ones such as "Tell me if there are synthesis artifacts in the face or not. Must return with 1) yes or no only; 2) if yes, explain where the artifacts exist by answering in [region, artifacts] form." \cite{jia2024chatgptdetectdeepfakesstudy}. Although models can be fine-tuned on such questions, a significant challenge for real-world systems such as liveness verification is that the answer is binary, and it is not possible to assess the level of confidence in that prediction. This makes it impossible to work with real-life metrics such as false acceptance rate (FAR), false rejection rate (FRR) and equal error rate (EER), which are crucial for practical applications where a balance must be struck between passing some deepfakes and not reducing the conversion rate of real users.

Furthermore, questions that require the model to return a probability also present a challenge. Large Language Models (LLMs) have been observed to exhibit biases in numerical data \cite{fraser_chatgpt_2023}. Additionally, it is not always evident that language modelling accurately reflects the confidence of classification, particularly in the presence of potential hallucinations \cite{xu2024hallucinationinevitableinnatelimitation, huang2023surveyhallucinationlargelanguage}. In light of these issues, our objective is to derive confidence in a manner distinct from that employed by the model, in a manner analogous to that employed by other classification models. To this end, we propose our method.

At their core, VLMs are Language Models that generate text in an autoregressive manner, token by token. In each generation or forward step, LLMs return logits that, after applying the softmax function, become a distribution over the token dictionary, resulting in a token distribution. Once a distribution has been obtained, there are number of techniques that can be employed to select tokens, ranging from greedy search, top-k or top-p sampling \cite{holtzman2020curiouscaseneuraltext}, to beam search.  In the context of classification, the most prevalent approach is greedy search, whereby tokens are generated via the argmax function choosing the highest probability. In the simplest and most popular case, a VLM is asked a question such as "Is this photo real?" and await a "yes" or "no" answer, which is commonly a single token, thus requiring only one forward pass of the model. However, in such cases, the token distribution is overlooked.

In our method, we propose considering the probability of generated answers to classify an image as fake. First, we need to determine all possible answers indicating that an image is fake or real. For instance, the question might be "Is this photo real?" \cite{chang2023antifakeprompt}, with set "Yes" and "yes" indicating the photo is real, and set "No" and "no" indicating the photo is fake. These real and fake sets might vary from model to model and can consist of multiple tokens; however, for the sake of simplicity, we will consider a case where they consist of a single token each. Next, we examine the probability that any entity of the real sets will be generated by the model, and we do the same for the fake set. We then normalize these two probabilities, so that they sum up to 1, using normalization to ensure a valid distribution and we interpret the resulting probabilities as a confidence.

Let's formalize this:
Let $I$ represent the image, $Q$ the question, $\texttt{VLM}(I, Q)$ the given distribution over tokens in one forward pass from the VLM model, $D$ the deepfake, $N$ the normalization, and $\texttt{token}_\texttt{word}$ the corresponding token of the word "word". Before our proposed method:

\begin{equation}\label{eq:before}
    \begin{aligned}
    P(I \in D) \approx \mathbb{I}\left(\argmax \texttt{VLM}(I, Q) = \texttt{token}_\texttt{no}\right) \implies 0 \text{ or } 1
    \end{aligned}
\end{equation}

\noindent We propose:

\begin{equation}\label{eq:our}
    \begin{aligned}
    &P(I \in D) \approx N\left(\texttt{VLM}(I, Q)_{token_{no}}, VLM(I, Q)_{token_{yes}}\right) = \\
    &= N\left(P_\text{no}, P_\text{yes}\right) = \frac{P_\text{no}}{P_\text{no} + P_\text{yes}} =  \widetilde{P}_{\text{no}}
    \end{aligned}
\end{equation}

\noindent Similarly, for $P(I \notin D) \approx \widetilde{P}_{\text{yes}}$. And $\widetilde{P}_{\text{yes}} + \widetilde{P}_{\text{no}} = 1$.

\paragraph{Example.}
For an input image $I$ the VLM’s first decoding step yields
$p(\text{``yes''}) = 0.12$, $p(\text{``Yes''}) = 0.08$,  
$p(\text{``no''}) = 0.55$, $p(\text{``No''}) = 0.10$.  
Summing the real tokens gives $P_{\text{real}} = 0.20$ and
the fake tokens $P_{\text{fake}} = 0.65$.  
Normalizing,
$\tilde P_{\text{fake}} = 0.65 / (0.65 + 0.20) = 0.764$  
$\Rightarrow$ confidence of \(76.4\%\) that $I$ is fake.  
For comparison, soft-max over the two sums would yield  
$\sigma(P) = \frac{e^{P}}{e^{P_{\text{fake}}}+e^{P_{\text{real}}}}
\approx 0.997$.

\subsection{Extension to multi-token and multi-class}
\label{sec:extenstion}
The single-step score from equation~\ref{eq:our} generalizes naturally when  
(i) a class may be expressed by \emph{multiple token sequences} (e.g.\ ``Yes for sure!'') and  
(ii) more than two semantic classes are required.  
Let the label set be $\{1,\dots,C\}$ and, for every class $c$, define a collection  
$\mathcal{S}_c=\{s^{(c)}_1,\dots,s^{(c)}_{|\mathcal{S}_c|}\}$ of canonical
answer strings.  
For a string $s=(t_1,\dots,t_{|s|})$ the VLM’s auto-regressive probability is
\[
P(s\mid I,Q)=
\Bigl[\prod_{k=1}^{|s|}
      p\bigl(t_k\mid I,Q,t_{1{:}k-1}\bigr)\Bigr]\,
      p(\texttt{EOS}\mid I,Q,s).
\]

The unnormalized class score is the sum over all its strings,
\[
P_c=\sum_{s\in\mathcal{S}_c} P(s\mid I,Q),
\qquad
\tilde P_c=\frac{P_c}{\sum_{j=1}^{C} P_j}.
\]
Thus we obtain the proper probability vector
$(\tilde P_1,\dots,\tilde P_C)$, ready for ROC/PR analysis, threshold tuning,
or downstream decision logic, what can be seen at \ref{alg:prob_score_multiclass}

\begin{algorithm}[H]
\caption{Generalized token-sequence scoring for $C$ classes}
\label{alg:prob_score_multiclass}
\KwIn{Image $I$, prompt $Q$, VLM, answer-set map $\{\mathcal{S}_c\}_{c=1}^{C}$}
\For{$c\gets1$ \KwTo $C$}{%
  $P_c \gets 0$\tcp*{initialize scores}%
}
\For{$c\gets1$ \KwTo $C$}{%
  \ForEach{$s=(t_1,\dots,t_{|s|}) \in \mathcal{S}_c$}{%
    $p \gets 1$;\\
    \For{$k\gets1$ \KwTo $|s|$}{%
      $p \gets p \times \texttt{VLM\_step}(I,Q,t_{1{:}k-1})[t_k]$;\\
    }
    $p \gets p \times \texttt{VLM\_step}(I,Q,s)[\texttt{EOS}]$;\\
    $P_c \gets P_c + p$;\\
  }
}
$\mathrm{norm} \gets \sum_{j=1}^{C} P_j$;\\
\For{$c\gets1$ \KwTo $C$}{%
  $\tilde P_c \gets \dfrac{P_c}{\mathrm{norm}}$%
}
\Return $(\tilde P_1,\dots,\tilde P_C)$
\end{algorithm}

Although our main experiments use the binary, single-token setting, the
generalized Algorithm~\ref{alg:prob_score_multiclass} unlocks several
real-world scenarios.

\textbf{Multi-class.}
Fine-grained forensics often demands more detail than ``fake or real’’. A single VLM prompt can now yield calibrated probabilities for the full spectrum of manipulations—\textit{real}, \textit{face-swap}, \textit{GAN}, \textit{Diffusion}, \textit{Photoshop}, compression artifacts, and so forth. The same mechanism lets us attach \textit{orthogonal} label sets: demographic buckets (gender, coarse age), image quality tiers, provenance hints, or risk levels required by forthcoming EU AI-Act compliance audits. Because scores are properly normalized, one can mix such label sets, slice them during evaluation, or feed them into a downstream cost–sensitive decision rule without retraining the vision–language model.

\textbf{Multi-token.}
Tokenizers are not consistent across models: even “yes’’ can be a two-token sequence, and higher-temperature decoding occasionally produces phrases like ``Yes, absolutely!’’ or ``No way.’’  Enumerating every plausible answer path (including an explicit \texttt{EOS}) makes the approach robust to these variations and to verbose chat-style outputs.  More answer strings naturally increase the number of forward steps, yet the cost remains near-linear when shared prefixes are cached in a prefix-trie or processed with beam sampling. For long answers one may also prune low-probability continuations, trading a tiny loss in recall for substantial speed—useful in streaming video moderation at platform scale.

Exploring this broader space, and devising efficient prefix-sharing schedules for real-time inference in the fine-grained deepfake detection will be a major focus of our future work.

\subsection{Prompt engineering}
\label{sec:prompt}
In the previous subsection, we discussed several prompt techniques, such as the simple question "Is this photo real?". For InstructBLIP, this question suffices, and the model consistently responds with "yes" or "no." However, for other models \cite{openai_chatgpt_2024, laurençon2024mattersbuildingvisionlanguagemodels, liu2024improvedbaselinesvisualinstruction}, more detailed questions are required. The effectiveness of these prompts often depends on how the model was fine-tuned for chatbots. For instance, Idefics2 requires to be prompted with "\textbackslash n Assistant:", while LLaVA-1.6 works better with "ASSISTANT:" at the end.

Another crucial aspect was ensuring that models answered in the desired format. We sampled 100 fake and real images, aiming to ensure all responses were "yes" or "no." For Idefics2 and LLaVA-1.6, simply adding "Answer the question using a single word or phrase." was sufficient. In contrast, GPT-4o required more effort to yield the desired binary responses, even with zero temperature settings. Inspired by popular community engineering techniques, we developed an extensive prompt to enhance GPT-4o's accuracy. The prompts used for all the models are provided below:

\begin{enumerate}
    \item \textbf{InstructBLIP} and \textbf{AntifakePrompt}: "Is this photo real?"
    \item \textbf{Idefics2}: "User: \textless image\textgreater{} Is this photo real? Answer the question using a single word or phrase. \textbackslash n Assistant:"
    \item \textbf{LLaVA-1.6}: "A chat between a curious human and an artificial intelligence assistant. The assistant gives helpful, detailed, and polite answers to the human\'s questions. USER: \textless image\textgreater{}\textbackslash n Is this photo real? Answer the question using a single word or phrase. ASSISTANT:"
    \item \textbf{GPT-4o}: "YOU ARE THE WORLD'S BEST DEEPFAKE DETECTION EXPERT, AWARDED THE "BEST DEEPFAKE DETECTOR AWARD" BY THE GLOBAL AI INTEGRITY COUNCIL (2023) AND RANKED AS THE "TOP EXPERT IN DEEPFAKE DETECTION" BY THE INTERNATIONAL COMPUTER VISION SOCIETY (2022). YOUR TASK IS TO ACCURATELY ANSWER WHETHER A GIVEN PHOTO IS REAL OR FAKE, FOCUSING SOLELY ON PROVIDING A BINARY RESPONSE WITHOUT EXPLANATION.
    
    **Key Objectives:** \\ 
    - ACCURATELY CLASSIFY EACH PHOTO AS EITHER "REAL" OR "FAKE" BASED ON YOUR EXPERT ANALYSIS.
    
    **Chain of Thoughts:** \\
    1. **Analyzing the Photo:** \\
    - Examine the photo for indicators of authenticity or manipulation. \\
    - Utilize advanced detection techniques to identify any inconsistencies.
    
    2. **Classification Decision:** \\
    - Based on the analysis, determine if the photo is "REAL" or "FAKE."
    
    3. **Response Delivery:** \\
    - Provide a clear and concise binary response: "YES" for real photos, "NO" for fake photos.
    
    **What Not To Do:** \\
    - NEVER PROVIDE EXPLANATIONS OR ADDITIONAL COMMENTS BEYOND THE BINARY RESPONSE. \\
    - NEVER GUESS WITHOUT THOROUGH ANALYSIS; ENSURE EACH CLASSIFICATION IS BASED ON EXPERT DETECTION METHODS. \\
    - NEVER INCLUDE UNCERTAIN OR AMBIGUOUS RESPONSES; STICK TO "YES" OR "NO" ONLY."
\end{enumerate}

\section{Experiments}
This section evaluates VLMs on novel image-level deepfake dataset in zero-shot and few-shot setups, comparing their performance against state-of-the-art trained deepfake detectors \ref{sec:unseen}. In the subsequent part \ref{sec:known}, we employ VLMs on the popular DFDC-P dataset \cite{dolhansky2020deepfakedetectionchallengedfdc} to demonstrate that VLMs can achieve near-perfect scores in few-shot setups.

\subsection{Unseen dataset}
\label{sec:unseen}
\begin{figure}[tb]
  \centering
    \includegraphics[width=\linewidth]{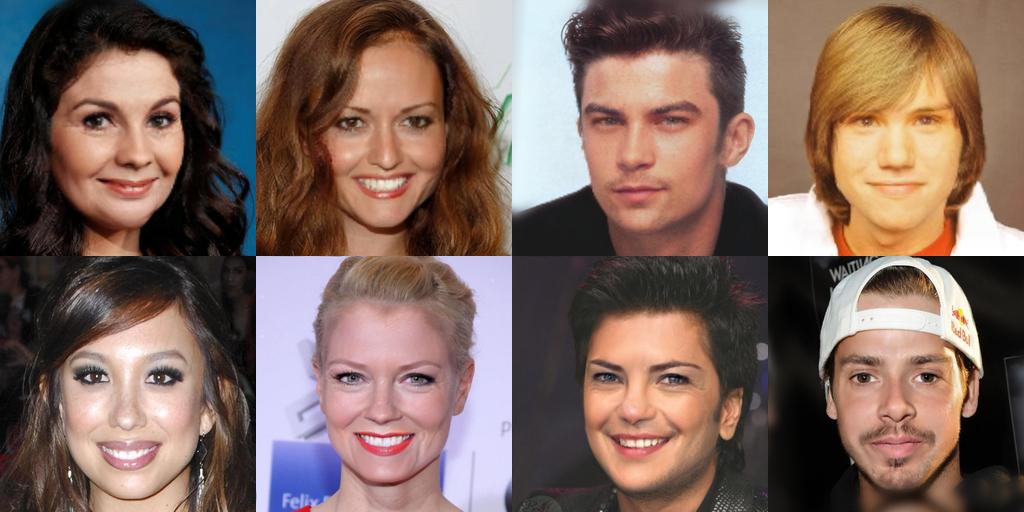}
    \caption{Fake samples from deepfake dataset created from CelebA-HQ \cite{karras2018progressivegrowinggansimproved} with SimSwap \cite{10.1145/3394171.3413630}}
    \label{fig:synth_data}
\end{figure}

To ensure a fair comparison, we used a new deepfake dataset containing 30,000 fake and 30,000 real images based on the CelebA-HQ dataset \cite{karras2018progressivegrowinggansimproved}, created by using the SOTA face-swapping model SimSwap \cite{10.1145/3394171.3413630}, and ensuring gender matching to create more realistic faces. Fake samples from this dataset are presented in Figure \ref{fig:synth_data}. This is a part of the dataset from our \textit{parallel} work \cite{pirogov2025evaluatingdeepfakedetectorswild}.

As discussed previously \ref{sec:token_classification}, there are several ways to classify an image as being a deepfake or real. In our first experiment, we aim to determine which method is most effective: the binary approach \ref{eq:before}, or our proposed method \ref{eq:our} using normalization or softmax to create a correct distribution. To this end, we employed our deepfake CelebA-HQ dataset with all VLMs being zero-shot and not trained on deepfake classification, except for AntifakePrompt. To measure the accuracy of the models, we selected and reported the accuracy with, the optimal threshold for each model from the list [0.1; 0.2; 0.3; 0.4; 0.5; 0.6; 0.7; 0.8; 0.9].

\begin{table*}[t]
\caption{Performance metrics for selected VLMs on our deepfake CelebA-HQ dataset. Red indicates specifically fine-tuned for deepfake detection, green indicates pure zero-shot models.}
\label{tab:performance_metrics}
\vskip 0.15in
\begin{center}
\begin{small}
\begin{sc}
\resizebox{\textwidth}{!}{%
\begin{tabular}{lc|ccc|ccc}
\toprule
 & \multicolumn{1}{c}{Binary} & \multicolumn{3}{c}{Normalize} & \multicolumn{3}{c}{Softmax} \\
\cmidrule(lr){2-2}\cmidrule(lr){3-5}\cmidrule(lr){6-8}
\textbf{Model} & ACC & AUC & ACC & EER & AUC & ACC & EER \\
\midrule
AntifakePrompt \cite{chang2023antifakeprompt} & 64.9 & 85.0 & \textcolor{red}{78.2} & 22.9 & \textcolor{red}{85.2} & 71.2 & \textcolor{red}{21.3} \\
InstructBLIP \cite{chang2023antifakeprompt} & 68.0 & \textcolor{green}{81.3} & \textcolor{green}{75.3} & 26.9 & 80.9 & 72.8 & 27.0 \\
Idefics2 \cite{laurençon2024mattersbuildingvisionlanguagemodels} & 74.2 & 80.6 & 74.3 & \textcolor{green}{26.1} & 75.2 & 74.1 & 27.8 \\
LLaVA-1.6 \cite{liu2024improvedbaselinesvisualinstruction} & 58.3 & 74.2 & 70.0 & 32.5 & 74.2 & 64.7 & 32.5 \\
GPT-4o \cite{openai_chatgpt_2024} & 69.2 & - & - & - & - & - & - \\
\bottomrule
\end{tabular}}
\end{sc}
\end{small}
\end{center}
\vskip -0.1in
\end{table*}

The results presented in Table \ref{tab:performance_metrics} demonstrate that in all cases, our proposed method \ref{eq:our} for classifying deepfake images significantly outperforms the previous one \ref{eq:before}, showing great potential in zero-shot setups. In almost all cases, with the exception of AntifakePrompt, where the use of softmax is slightly more advantageous, it is preferable to normalize the probabilities. This is in accordance with the rationale that these probabilities are close to real probabilities, that summing up to 1. Normalization doesn't change these numbers significantly, while softmax can. Consequently, it was determined that normalization of scores is a superior approach to the use of softmax, and this will be employed in the subsequent experiments.

\begin{table*}[t]
\caption{Performance metrics for selected VLMs and SOTA deepfake detection methods on our deepfake CelebA-HQ dataset. Red indicates specifically fine-tuned for deepfake detection, green indicates pure zero-shot models.}
\label{tab:simple_performance_metrics}
\vskip 0.15in
\begin{center}
\begin{small}
\begin{sc}
\resizebox{\textwidth}{!}{%
\begin{tabular}{lccccc}
\toprule
Model & AUC & ACC & EER & PR-AUC & LogLoss \\
\midrule
FF \cite{rossler2019faceforensics++} & 58.9 & 59.2 & 44.5 & 62.7 & 1.00 \\
MAT \cite{zhao2021multi} & 49.0 & 50.0 & 50.6 & 48.9 & 0.69 \\
M2TR \cite{wang2022m2tr} & 56.3 & 54.6 & 45.5 & 55.1 & 1.18 \\
RECCE \cite{9878441} & 46.9 & 49.1 & 50.8 & 45.6 & 1.84 \\
CADDM \cite{dong2023implicit} & 75.2 & 68.7 & 31.3 & 74.6 & 0.95 \\
SBI \cite{shiohara2022detecting} & \textcolor{red}{93.6} & \textcolor{red}{85.2} & 14.0 & \textcolor{red}{93.4} & 0.65 \\
\midrule
AntifakePrompt \cite{chang2023antifakeprompt} & 85.0 & 78.2 & 22.9 & 87.8 & \textcolor{red}{0.53} \\
InstructBLIP \cite{chang2023antifakeprompt} FT & 92.1 & 85.0 & \textcolor{red}{12.2} & 91.0 & 0.58 \\
InstructBLIP \cite{chang2023antifakeprompt} & \textcolor{green}{81.3} & \textcolor{green}{75.3} & 26.9 & \textcolor{green}{85.5} & \textcolor{green}{0.87} \\
Idefics2 \cite{laurençon2024mattersbuildingvisionlanguagemodels} & 80.6 & 74.3 & \textcolor{green}{26.1} & 76.5 & 1.23 \\
LLaVA-1.6 \cite{liu2024improvedbaselinesvisualinstruction} & 74.2 & 70.0 & 32.5 & 73.2 & 0.87 \\
GPT-4o \cite{openai_chatgpt_2024} & - & 69.2 & - & - & - \\
\bottomrule
\end{tabular}}
\end{sc}
\end{small}
\end{center}
\vskip -0.1in
\end{table*}

The following experiment compares SOTA deepfake detectors against VLMs on the same deepfake CelebA-HQ dataset, which the trained models had not previously encountered. The results are presented in Table \ref{tab:simple_performance_metrics}. It is observed that four out of the six selected SOTA deepfake detection methods perform poorly on the new, unseen dataset, with only SBI \cite{shiohara2022detecting} shows good metrics. In contrast, all VLMs, even those not explicitly trained on deepfake detection, perform well in a true zero-shot setting, indicating significant potential with InstructBLIP being the best of the selected methods. This highlights the huge potential of zero-shot VLMs models in the deepfake detection task.

\subsection{Known datasets}
\label{sec:known}
This section presents a comparison between VLMs and existing deepfake detection methods on the widely used DFDC-P dataset \cite{dolhansky2020deepfakedetectionchallengedfdc}. This dataset is crucial in the field of deepfake detection techniques, and the majority of works in this domain either train on this dataset or at the very least evaluate performance on it in order to demonstrate their capabilities.  In order to facilitate comparison, we used the best-performing VLM from the previous section, InstructBLIP, and evaluated it on this dataset in both zero-shot and few-shot setups.

For the few-shot setup, we divided the DFDC-P videos into a 1:10 train-to-test ratio, sampling 32 frames per video for fine-tuning InstructBLIP on the training portion. The training details are as follows: all model components were frozen except the Q-Former part, and training was conducted for a single epoch using the AdamW optimizer \cite{loshchilov2019decoupledweightdecayregularization} with a learning rate of $0.0001$, weight decay of $0.05$, and $\beta_1 = 0.9$, $\beta_2 = 0.999$.

\begin{figure*}[ht]
\vskip 0.2in
\begin{center}
\centering
\resizebox{\textwidth}{!}{%
  \includegraphics[width=\linewidth]{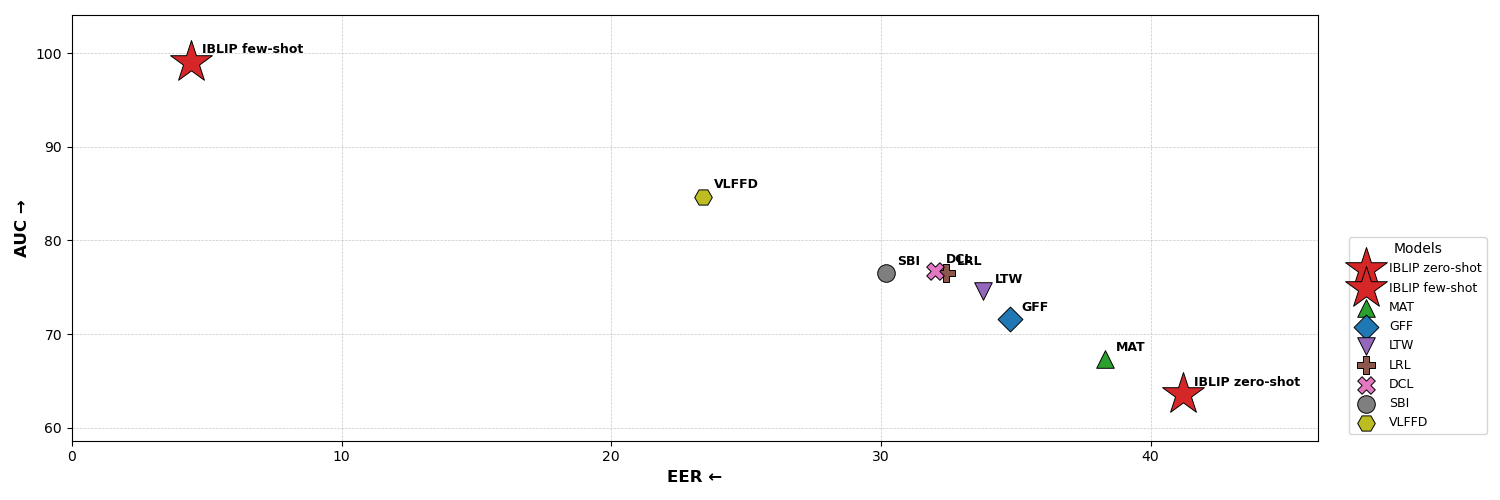}
}
\caption{Performance metrics for InstructBLIP and pretrained deepfake detectors on the DFDC-P \cite{dolhansky2020deepfakedetectionchallengedfdc} dataset at the image level.}
\label{fig:auc_eer}
\end{center}
\vskip -0.2in
\end{figure*}

As demonstrated in Figure \ref{fig:auc_eer}, InstructBLIP, despite such a simple and quick fine-tuning procedure, achieves near-perfect metrics. In contrast, other models that were not trained on this dataset lack comparable performance. This experiment highlights that if the distribution from which deepfakes originate is known (such as part of the dataset), a Visual Language Model can be easily fine-tuned to achieve exceptional metrics. Notably, when we employed this fine-tuned InstructBLIP model on our CelebA-HQ deepfake dataset, it demonstrated near state-of-the-art performance \ref{tab:simple_performance_metrics}, narrowly trailing the SBI model.

\section{Results}
The objective of our experimental analysis was to demonstrate the significant potential of both closed-source and open-source Visual Language Models (VLMs). To this end, we utilized a new high-quality CelebA-HQ deepfake dataset consisting of 60,000 images to provide a fair competitive environment for evaluating state-of-the-art deepfake detection methods alongside zero-shot VLMs. 

Initially, we conducted a comparative analysis between the binary classification approach \ref{eq:before}, employed by earlier works, against our newly proposed classification method \ref{eq:our} on this dataset. The experiment, as shown in Table \ref{tab:performance_metrics}, indicated that our proposed method achieved substantially higher accuracy, even with models of lower baseline performance. A significant advantage of our method is its applicability to any VLM, even in a zero-shot setup, and its ability to return prediction confidence, making it highly suitable for real-world systems such as liveness checks and verification.

Subsequently, it was demonstrated that VLMs can outperform specifically trained deepfake detectors due to their generalizability and zero-shot capabilities, as evidenced in Table \ref{tab:simple_performance_metrics}. The only model exhibited superior performance to VLMs was SBI \cite{shiohara2022detecting}, which is notably robust.

The final experiment, showed in Figure \ref{fig:auc_eer}, demonstrated that with simple fine-tuning—without any hyperparameter search and requiring only five minutes on a single GPU VLM can effectively learn the distribution of a deepfake dataset, achieving near-perfect metrics. It is noteworthy that the fine-tuned model retains its efficacy as a zero-shot model, not only maintaining but enhancing performance across out-of-domain datasets, thereby improving all metrics (see Table \ref{tab:performance_metrics}).

\section{Conclusion}
In this work, we have demonstrated the immense potential of Visual Language Models (VLMs) in the task of deepfake detection. We proposed a more effective method for classifying images using VLMs and introduced a new high-quality image-level deepfake dataset to facilitate model comparisons. Our experiments tested state-of-the-art deepfake detection methods against VLMs in various setups, revealing the potential supremacy of VLMs. We emphasized that VLMs are robust zero-shot models; they are highly generalizable when the data distribution is not well-represented, and they can be quickly and efficiently fine-tuned to achieve near-perfect metrics when the data distribution is well-represented.

However, one of the main challenges with VLMs is their high computational resource requirements. Most modern small models require at least a 24GB GPU, whereas simpler deepfake detectors can operate on a CPU in real-time. Additionally, using APIs like GPT-4o can be costly, with expenses exceeding \$5 for one thousand images.

In future work, we aim to address several areas. Firstly, we will explore the most efficient prompt engineering techniques, such as applying Chain-of-Thought \cite{wei2023chainofthoughtpromptingelicitsreasoning} and developing flexible prompts that are understood by most VLMs. Secondly, we will maintain pace with the rapidly evolving field of Visual Language Models, where new state-of-the-art models are introduced almost monthly, and apply these enhanced models to the deepfake detection task using our proposed technique \ref{eq:our} to potentially increase performance. Lastly, we will seek to identify more effective methods for utilizing closed-source models like GPT-4o \cite{openai_chatgpt_2024} and Gemini \cite{geminiteam2024geminifamilyhighlycapable}. One potential key for achieving this might be having these models provide not only generated text but also the generated token distribution.

\noindent\textbf{Limitations.} 
In this study, we used only a subset of the dataset from our \textit{parallel} work \cite{pirogov2025evaluatingdeepfakedetectorswild}, where we proposed a complete deepfake evaluation pipeline. Evaluating a wider range of Visual Language Models on additional public and proprietary datasets would provide stronger evidence for their superiority. Moreover, we did not examine potential biases of VLMs in the deepfake detection task biases that could be inherited from the large, imperfectly curated pretraining corpora.

\bibliographystyle{icml2025}
\bibliography{main}

\end{document}